\documentclass{article} % For LaTeX2e
\usepackage{iclr2027_conference,times}

\usepackage{amsmath}
\usepackage{amssymb}
\usepackage{array}
\usepackage{booktabs}
\usepackage{colortbl}
\usepackage{enumitem}
\usepackage{float}
\usepackage{placeins}
\usepackage{graphicx}
\usepackage{caption}
\usepackage{subcaption}
\usepackage{tabularx}
\usepackage{mathtools}
\usepackage{microtype}
\usepackage{multirow}
\usepackage{xcolor}
\definecolor{gljepablue}{RGB}{235,245,252}
\definecolor{lowergray}{RGB}{145,145,145}
\usepackage{wrapfig}
\usepackage{hyperref}
\usepackage{url}

\newcommand{\method}{\mbox{GL-JEPA}}
\newcolumntype{Y}{>{\centering\arraybackslash}X}
\newcolumntype{L}[1]{>{\raggedright\arraybackslash}p{#1}}

\newcolumntype{P}{
  >{\centering\arraybackslash
    \hsize=.9\hsize
  \linewidth=\hsize}X
}

\newcolumntype{E}{
  >{\centering\arraybackslash
    \hsize=1.35\hsize
  \linewidth=\hsize}X
}

\title{Learning Graph Representations via Predictive Language Supervision}

\author{Jongwoo Kim, Chanjae Song, Jeongwhan Choi, Woosung Kang, Noseong Park \& Mun Yong Yi\thanks{Corresponding author.} \\
  Korea Advanced Institute of Science and Technology (KAIST), Daejeon, South Korea \\
  \texttt{\{gsds4885,munyi\}@kaist.ac.kr}
}

\iclrfinalcopy

\begin{document}

\maketitle

\begin{abstract}
Graph--language pretraining typically relies on autoregressive token prediction or graph--text alignment. Token prediction ties supervision to a particular \emph{linguistic realization}, while alignment may suppress information that is unnecessary for matching but useful for downstream graph tasks. We introduce \textbf{GL-JEPA}, which uses a frozen Text Encoder to define a continuous target for each description and a Predictor to reconstruct this target from graph states. Predicting the full description latent reduces dependence on surface form while encouraging graph states to capture description information that may otherwise be suppressed by alignment. For downstream adaptation, query conditioning allows the Predictor to extract task-relevant information from frozen graph representations. Compared with token prediction, GL-JEPA improves by 25.2--43.8 percentage points the rate at which two valid descriptions both outrank a negative. In a controlled diagnostic, latent prediction achieves 95.6\% probe accuracy on a target-relevant factor, versus 12.7\% for alignment with easy negatives. GL-JEPA also outperforms direct alignment on node-, link-, and graph-level tasks and supports query-conditioned graph question answering across nine tasks with a frozen Graph Encoder.

\end{abstract}

\section{Introduction}
\label{sec:introduction}

\begin{wrapfigure}[21]{r}{0.46\textwidth}
  \centering
  \vspace{-4mm}

  \begin{subfigure}[t]{\linewidth}
    \centering
    \includegraphics[
      width=0.98\linewidth,
      trim=0 0 0 0mm,
      clip
    ]{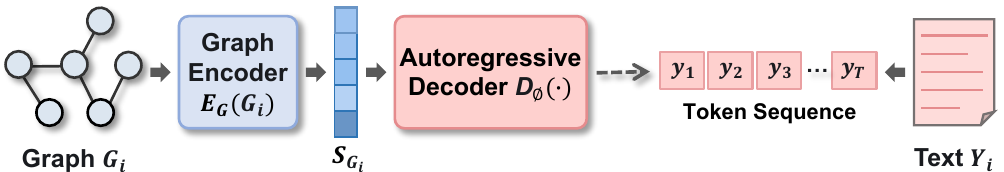}
    \caption{Autoregressive Token Prediction}
    \label{fig:gljepa_overview_a}
  \end{subfigure}
  \vspace{1mm}

  \begin{subfigure}[t]{\linewidth}
    \centering
    \includegraphics[
      width=0.98\linewidth,
      trim=0 0 0 0mm,
      clip
    ]{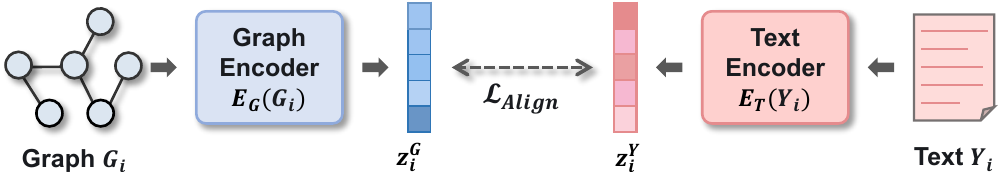}
    \caption{Graph--Text Alignment}
    \label{fig:gljepa_overview_b}
  \end{subfigure}
  \vspace{1mm}

  \begin{subfigure}[t]{\linewidth}
    \centering
    \includegraphics[
      width=0.98\linewidth,
      trim=0 0 0 0mm,
      clip
    ]{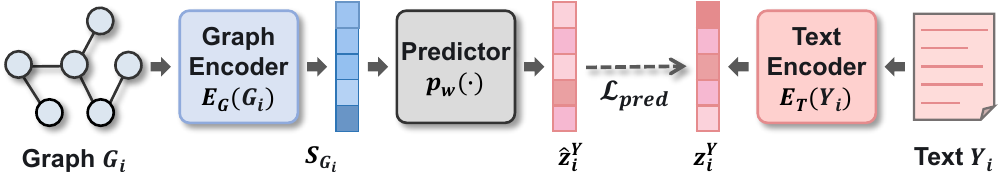}
    \caption{Graph--Language Latent Prediction (Ours)}
    \label{fig:gljepa_overview_c}
  \end{subfigure}
  \vspace{-2mm}
  \caption{
    Graph--language supervision interfaces.
    Token prediction reconstructs an observed wording.
    Direct alignment matches paired graph and description representations.
    GL-JEPA uses a separate Predictor to infer the frozen description target
    from graph context.
  }
  \label{fig:gljepa_overview}
\end{wrapfigure}

Graph--language models learn representations of entities and relations by integrating relational structure with language \citep{plenz2024glm,jin2024llmongraphs}. The graph input comprises directed structure, node-text attributes, and positional features, while the paired description is withheld from the Graph Encoder (\(E_G\)) and used only as language supervision. Unlike task-specific labels, which supervise predefined prediction targets, a paired description can capture multiple complementary aspects of a graph \citep{zhu2025graphclip,kong2025gofa}. Graph--language pretraining \textcolor{black}{leverages this richer supervision to learn transferable graph representations} for downstream tasks such as node classification, link prediction, graph--text retrieval, and graph question answering \citep{wen2023g2p2,zhu2025graphclip,tang2023graphgpt,kong2025gofa}.

Existing methods primarily use language supervision through two interfaces
(Figure~\ref{fig:gljepa_overview}).
Let \(G\) denote a graph with paired description \(Y\) and optional task query \(Q\).
\textbf{Autoregressive token prediction}
(Figure~\ref{fig:gljepa_overview}\subref{fig:gljepa_overview_a})
uses a language decoder \(D_\phi\), conditioned on \(E_G(G)\) and \(Q\), to generate \(Y\) as
\(D_\phi(E_G(G),Q) \rightarrow Y\)
\citep{tang2023graphgpt,chen2024llaga,kong2025gofa}.
Its loss is defined over the observed token sequence of \(Y\), thereby supervising both graph-relevant content and its particular \emph{linguistic realization}. The same graph semantics can be expressed through different entity orders, relation phrasings, sentence structures, lexical choices, and token orderings. Autoregressive training therefore incurs loss on surface-form variation that may be irrelevant to downstream graph tasks
\citep{tang2023graphgpt,chen2024llaga,kong2025gofa}.

\textbf{Direct graph--text alignment}
(Figure~\ref{fig:gljepa_overview}\subref{fig:gljepa_overview_b})
optimizes similarity between a graph embedding and the embedding of its paired description
\citep{wen2023g2p2,wang2023congrat,zhu2025graphclip,liu2026adaligner}.
Let \(E_T\) denote the Text Encoder for the paired description \(Y\).
With method-specific projections, alignment directly matches
\(E_G(G) \leftrightarrow E_T(Y)\).
This objective supervises the information needed to distinguish matched from mismatched graph--text pairs.
Features that do not improve the graph--text alignment objective may be suppressed even when useful for downstream tasks, creating an \emph{objective mismatch}.
Graphs may share entities and topics yet differ in relation direction or connectivity.
Matching constrains representations to information sufficient for graph--text correspondence. Downstream graph tasks may require additional structural distinctions
\citep{luo2024graphinstruct,kong2025gofa}.

Autoregressive token prediction uses the paired description as supervision but requires reconstructing a particular linguistic realization of \(Y\). Direct graph--text alignment avoids token-level reconstruction, but may weakly supervise features that are not necessary for graph--text matching. Neither existing interface directly optimizes for both properties.

\begin{center}
\textbf{\emph{\shortstack{
Can paired descriptions supervise graph representations\\
without token reconstruction while reducing feature suppression?
}}}

\end{center}

To address this question, we introduce \textbf{GL-JEPA}
(Figure~\ref{fig:gljepa_overview}\subref{fig:gljepa_overview_c}), which draws on
joint-embedding predictive architectures \citep{assran2023self,chen2025vljepa}
to predict a latent description target from graph context.
A frozen Text Encoder \(E_T\) encodes the paired description \(Y\) into a latent description target in a continuous embedding space.
The Graph Encoder \(E_G\) processes the graph structure and node-text attributes into contextualized node states and a graph summary. The Predictor \(P_\omega\) predicts the description target from these graph states as
\(P_\omega(E_G(G)) \rightarrow E_T(Y)\).
Training compares the predicted and target representations. The Graph Encoder \(E_G\) and Predictor \(P_\omega\) are updated while \(E_T\) remains frozen.
The paired description is used only to define the target and is not provided to the graph branch.

Because the description is predicted as a continuous latent representation rather than an observed token sequence, GL-JEPA avoids reconstructing a particular linguistic realization of (Y). Predicting the full description embedding encourages \(E_G\) to capture description-relevant graph features and may mitigate feature suppression. During downstream adaptation, \(P_\omega\) additionally conditions its readout on the task query \(Q\).

Our controlled experiments compare GL-JEPA with autoregressive token prediction and direct graph--text Alignment, and evaluate feature suppression using a synthetic diagnostic. Compared with token prediction, GL-JEPA achieves stronger language-prediction performance while reducing measured training-update cost by 66.0\%, including generation-readout fitting. Compared with Alignment, GL-JEPA improves Node accuracy by 3.59 points and Link AUC by 3.82 points, whereas Alignment performs better on bidirectional Retrieval R@1. In the feature-suppression diagnostic, a target-relevant factor is more recoverable from latent-prediction graph states than from Alignment graph states when matching does not require that factor. In complete-system frozen-transfer experiments, GL-JEPA attains the highest mean scores on Node, Link, and Graph tasks among the compared methods, while GraphCLIP performs best on Retrieval. With the 1B Graph Encoder frozen, GL-JEPA also achieves competitive performance across nine graph question-answering tasks.

Our contributions are as follows.
\begin{itemize}[leftmargin=*]

\item We identify two challenges in language supervision for text-attributed graphs: token reconstruction is tied to a particular linguistic realization, while direct graph--text alignment may suppress target-relevant features that are unnecessary for pair discrimination.

\item We introduce \textbf{GL-JEPA} for text-attributed graphs. A frozen Text Encoder defines a continuous target from the paired description, while the Graph Encoder and Predictor predict this target from graph structure and node text. This formulation avoids token-level reconstruction and \textcolor{black}{encourages retention of description-relevant features beyond those required for direct graph--text matching}.

\item We conduct controlled comparisons against autoregressive token prediction and direct graph--text alignment, together with a synthetic diagnostic for feature suppression. We further evaluate frozen transfer and query-conditioned graph question answering to assess the quality and usability of the learned graph representations.

\end{itemize}

\section{Method}
\label{sec:method}

\subsection{Problem Setup}
\label{sec:problem-definition}

Let \(\mathcal D=\{(G_i,Y_i)\}_{i=1}^{M}\) denote graphs paired with descriptions.
Each graph \(G_i=(V_i,E_i,\mathcal X_i)\) contains \(N_i=|V_i|\) nodes, directed edges \(E_i\), and node texts \(\mathcal X_i=\{X_{in}\}_{n=1}^{N_i}\).
The paired description \(Y_i\) provides supervision during pretraining and is not included in the graph input.
A frozen Text Encoder \(E_T\) maps each node text to \(\mathbf x_{in}=E_T(X_{in})\) and the paired description to a latent target \(\mathbf z_i^Y=E_T(Y_i)\in\mathbb R^{d_T}\).
The Graph Encoder \(E_G\) processes the graph structure and node features to produce contextualized node states \(\mathbf H_i\) and a graph summary \(\mathbf g_i\), which the Predictor \(P_\omega\) uses to predict \(\mathbf z_i^Y\).
During paired-description pretraining, \(E_G\) and \(P_\omega\) are updated while \(E_T\) remains fixed.

Graph Context Pretraining (Section~\ref{sec:structural-initialization}) initializes \(E_G\) by predicting target representations of masked nodes from a masked graph.
Graph--Language Latent Prediction (Section~\ref{sec:gljepa-architecture}) then trains \(E_G\) and \(P_\omega\) to predict the paired-description target \(\mathbf z_i^Y\).

\subsection{Graph Context Pretraining}
\label{sec:structural-initialization}

\paragraph{Context and target views.}
For each graph \(G_i\), we construct a masked context view
\(G_i^{\mathrm{ctx}}=\mathcal M(G_i)\)
and an unmasked target view
\(G_i^{\mathrm{tgt}}=G_i\).
Let \(\mathcal V_i^{\mathrm{mask}}\subseteq V_i\) denote the sampled nodes whose input features are masked.
\textcolor{black}{The masking operator \(\mathcal M\) preserves the node set, masks the features of nodes in
\(\mathcal V_i^{\mathrm{mask}}\), independently masks sampled feature coordinates, and drops sampled directed edges.}

The context view is processed by the online Graph Encoder \(E_G(\cdot;\theta)\).
The target view is processed by an encoder initialized from the online Graph Encoder.
Its parameters \(\bar\theta\) are updated by exponential moving average (EMA) as
\(\bar\theta \leftarrow \tau\bar\theta+(1-\tau)\theta\),
where \(\tau\) is the EMA momentum.

The two encoders produce
\(\widetilde{\mathbf H}_i=E_G(G_i^{\mathrm{ctx}};\theta)\)
and
\(\mathbf H_i^+=\operatorname{sg}(E_G(G_i^{\mathrm{tgt}};\bar\theta))\).
Here,
\(\widetilde{\mathbf H}_i=
[\widetilde{\mathbf h}_{i1};\ldots;\widetilde{\mathbf h}_{iN_i}]\)
and
\(\mathbf H_i^+=[
\mathbf h_{i1}^+;\ldots;\mathbf h_{iN_i}^+]\).
The vectors \(\widetilde{\mathbf h}_{in}\) and \(\mathbf h_{in}^+\) are the representations of the same node \(n\) from the masked and unmasked views, respectively.
The stop-gradient operator \(\operatorname{sg}\) blocks gradients through the target encoder.
Appendix~\ref{app:implementation-details} reports the node-masking rate, edge-drop rate, EMA schedule, and optimization settings.

\paragraph{Masked-node prediction and objective.}
The structural Predictor \(P_\psi\) is an MLP that predicts the target representation of each masked node as
\(\widehat{\mathbf h}_{in}=P_\psi(\widetilde{\mathbf h}_{in})\)
for \(n\in\mathcal V_i^{\mathrm{mask}}\).
Unmasked nodes provide context to the Graph Encoder but do not contribute directly to the prediction loss.

For a minibatch \(\mathcal B\), Graph Context Pretraining minimizes the normalized Smooth-L1 loss between the predicted and target representations of masked nodes as
\begin{equation}
\mathcal L_{\mathrm{GCP}}
=
\frac{1}{|\mathcal B|}
\sum_{i\in\mathcal B}
\frac{1}{|\mathcal V_i^{\mathrm{mask}}|}
\sum_{n\in\mathcal V_i^{\mathrm{mask}}}
\ell_{\mathrm{SL1}}
\left(
u_\varepsilon(\widehat{\mathbf h}_{in}),
u_\varepsilon(\mathbf h_{in}^+)
\right),
\label{eq:structural-loss}
\end{equation}
where
\(u_\varepsilon(\mathbf v)=\mathbf v/\max(\lVert\mathbf v\rVert_2,\varepsilon)\)
with \(\varepsilon=10^{-8}\), and \(\ell_{\mathrm{SL1}}\) denotes the coordinate-mean Smooth-L1 loss.
Gradients update the online Graph Encoder \(E_G\) and \(P_\psi\).
After Graph Context Pretraining, \(P_\psi\) and the target encoder are discarded, and the learned Graph Encoder initializes Graph--Language Latent Prediction.

\subsection{Graph--Language Latent Prediction}
\label{sec:learning-gljepa}
\label{sec:gljepa-architecture}

Graph--Language Latent Prediction uses the paired description \(Y_i\) to supervise the Graph Encoder.
The frozen Text Encoder \(E_T\) maps \(Y_i\) to the latent target \(\mathbf z_i^Y\), while \(E_G\) produces contextualized node states and a graph summary from \(G_i\).
The Predictor \(P_\omega\) predicts \(\mathbf z_i^Y\) from these graph representations.
Neither \(Y_i\) nor \(\mathbf z_i^Y\) is provided to \(E_G\) or \(P_\omega\).

\paragraph{Text Encoder (\(E_T\)).}
The frozen Text Encoder \(E_T\) maps each node text \(X_{in}\) to
\(\mathbf x_{in}=E_T(X_{in})\)
and the paired description \(Y_i\) to
\(\mathbf z_i^Y=E_T(Y_i)\in\mathbb R^{d_T}\).
The same encoder and pooling rule are used for both inputs.
The node features \(\{\mathbf x_{in}\}_{n=1}^{N_i}\) are supplied to \(E_G\),
whereas \(\mathbf z_i^Y\) is used only as the prediction target.
\(E_T\) remains frozen throughout training.

\begin{wrapfigure}{r}{0.35\textwidth}
 \centering
 \vspace{-2mm}
 \includegraphics[width=\linewidth]{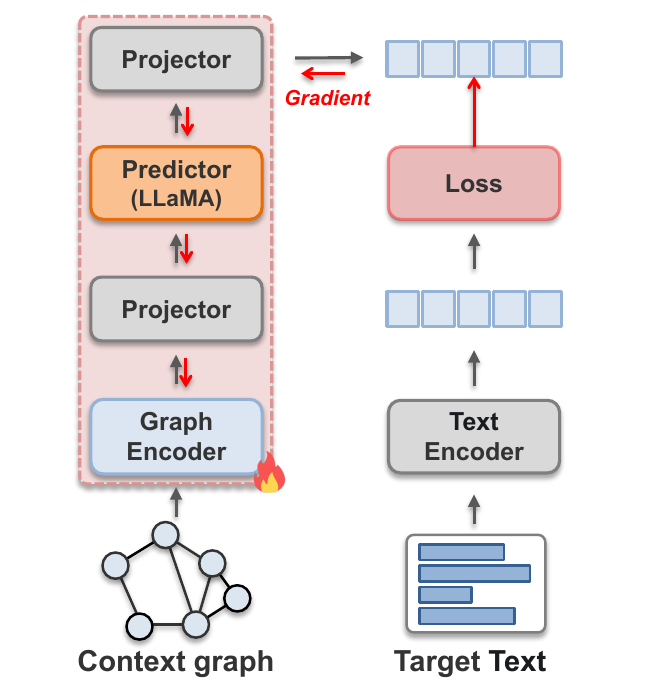}

\caption{Graph--Language Latent Prediction.
  The frozen target \(\mathbf z_i^Y\) is not provided to \(E_G\) or \(P_\omega\).
  Only the Graph Encoder and Predictor are trained.}
 \label{fig:latent-prediction-stage}
 \vspace{-5mm}
\end{wrapfigure}

\paragraph{Graph Encoder (\(E_G\)).}
The Graph Encoder takes the directed graph structure and node features
\(\{\mathbf x_{in}\}_{n=1}^{N_i}\) as input.
We instantiate \(E_G\) with GraphGPS \citep{rampasek2023graphgps},
using GraphSAGE \citep{hamilton2017inductive} for local message passing
and self-attention for global mixing.
GraphSAGE aggregates information over directed edge neighborhoods,
while self-attention exchanges information across nodes.
The graph is not symmetrized and reverse edges are not added.

The Graph Encoder produces contextualized node states
\(\mathbf H_i=
[\mathbf h_{i1},\ldots,\mathbf h_{iN_i}]^\top
\in\mathbb R^{N_i\times d_G}\).
We compute the mean node state
\(\boldsymbol\mu_i=N_i^{-1}\sum_{n=1}^{N_i}\mathbf h_{in}\)
and obtain the graph summary as
\begin{equation}
\mathbf g_i
=
W_g\boldsymbol\mu_i+\mathbf b_g
\in\mathbb R^{d_G}.
\label{eq:graph-readout}
\end{equation}
The Predictor receives both \(\mathbf H_i\) and \(\mathbf g_i\).
Layer counts, state dimensions, positional features, and attention settings are reported in Appendix~\ref{app:implementation-details}.

\paragraph{Predictor (\(P_\omega\)).}
The Predictor maps the graph summary and contextualized node states to a prediction
\(\widehat{\mathbf z}_i^Y\) of the latent target.
The graph summary and node states are projected from dimension \(d_G\) to
the Predictor dimension \(d_P=2048\).
Let
\(a_G(\mathbf v)=W_G\mathbf v+\mathbf b_G\)
and
\(a_N(\mathbf v)=W_N\mathbf v+\mathbf b_N\),
where
\(W_G,W_N\in\mathbb R^{d_P\times d_G}\).
Learned embeddings
\(\mathbf e_G,\mathbf e_N\in\mathbb R^{d_P}\)
identify the graph-summary and node tokens.
The Predictor input sequence is
\begin{equation}
\mathbf C_i
=
\left[
a_G(\mathbf g_i)+\mathbf e_G;
\left(a_N(\mathbf h_{in})+\mathbf e_N\right)_{n=1}^{N_i}
\right]
\in\mathbb R^{(N_i+1)\times d_P}.
\label{eq:graph-token-sequence}
\end{equation}

We initialize \(P_\omega\) from the final eight Llama-3.2-1B Transformer blocks
\citep{meta2024llama32} and disable causal masking.
The Transformer processes all valid graph-derived tokens bidirectionally.
\textcolor{black}{The Predictor receives node states already contextualized by graph structure and their order-independent mean summary.}
The Predictor assigns consecutive rotary positions to node tokens.

For a minibatch, let
\(L=1+\max_iN_i\),
\(\widetilde{\mathbf C}_i\in\mathbb R^{L\times d_P}\)
be the padded Predictor input,
and \(m_{ij}\in\{0,1\}\) indicate a valid position.
Let
\(\mathbf U_i=\mathcal T_\omega(\widetilde{\mathbf C}_i;m_i)\)
be the Transformer output and
\(\mathbf u_{ij}=\mathbf U_i[j]\)
the output at position \(j\).
The Predictor averages the valid outputs and projects them to the Text Encoder dimension as
\begin{equation}
\widehat{\mathbf z}_i^Y
=
q_\omega\!\left(
\frac{\sum_{j=0}^{L-1}m_{ij}\mathbf u_{ij}}
     {\sum_{j=0}^{L-1}m_{ij}}
\right),
\qquad
q_\omega:\mathbb R^{d_P}\rightarrow\mathbb R^{d_T}.
\label{eq:predictor-readout}
\end{equation}
Padding positions are excluded from attention and pooling.
The input projections, type embeddings, Transformer blocks, and output projection are trainable parts of \(P_\omega\).

\paragraph{Training objective.}
For each pair \((G_i,Y_i)\), the Predictor produces
\(\widehat{\mathbf z}_i^Y=P_\omega(\mathbf H_i,\mathbf g_i)\),
while the frozen Text Encoder produces
\(\mathbf z_i^Y=E_T(Y_i)\).
The predicted and target representations are compared only through the prediction loss.
We normalize each representation as
\(u_\varepsilon(\mathbf v)=
\mathbf v/\max(\lVert\mathbf v\rVert_2,\varepsilon)\),
with \(\varepsilon=10^{-8}\), and minimize
\begin{equation}
\mathcal L_{\mathrm{GL\mbox{-}JEPA}}
=
\frac{1}{|\mathcal B|d_T}
\sum_{i\in\mathcal B}
\left\|
u_\varepsilon(\widehat{\mathbf z}_i^Y)
-
u_\varepsilon(\mathbf z_i^Y)
\right\|_2^2.
\label{eq:gljepa-loss}
\end{equation}
The normalization removes representation magnitude from the loss, and
\(1/d_T\) averages the squared error over target coordinates.
Gradients update \(E_G\) and \(P_\omega\), while \(E_T\) remains frozen.
The objective uses neither token likelihood nor in-batch negatives.

\section{Experiments}
\label{sec:experiments}

\textcolor{black}{Section~\ref{sec:rq1-controlled} compares supervision interfaces. Section~\ref{sec:continuous-supervision} analyzes language prediction and computational cost, while Section~\ref{sec:feature-suppression} contrasts Alignment with easy and hard negatives against full-target latent prediction. Sections~\ref{sec:zsl-retrieval} and~\ref{sec:graph-question-answering} evaluate frozen transfer and graph question answering. Finally, Section~\ref{sec:ablation} examines the design choices of GL-JEPA.}

\begin{figure*}[t]
  \centering

  % =========================
  % Left: Table
  % =========================
  % =========================
  % Right: Figure
  % =========================
  \begin{minipage}[t]{\textwidth}
    \vspace{0pt}
    \centering

    \includegraphics[
      width=0.85\linewidth,
      keepaspectratio
    ]{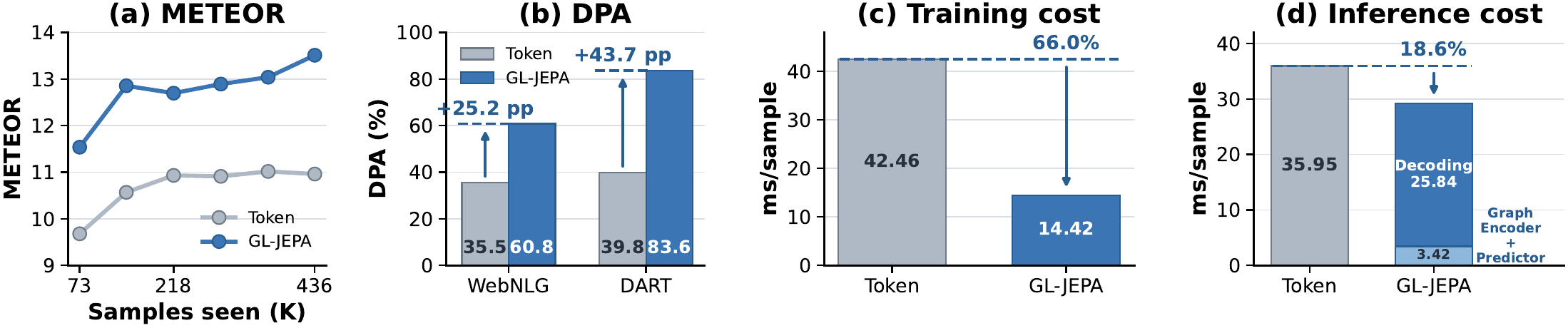}

    \vspace{-1mm}

\captionof{figure}{Token versus latent prediction. (a) WebNLG METEOR by pretraining samples (K). (b) Dual-positive accuracy on WebNLG factual negatives and DART mismatched descriptions. (c) Training-update latency, with GL-JEPA including separately measured generation-readout fitting and excluding decoding. (d) WebNLG inference latency, decomposed into graph encoding and prediction, and greedy decoding for GL-JEPA.}
\label{fig:dpa-efficiency}

  \end{minipage}
 \vspace{-3mm}
\end{figure*}

\subsection{Controlled Comparison of Graph--Language Supervision Interfaces}
\label{sec:rq1-controlled}

\paragraph{Setup.}
Table~\ref{tab:frozen-transfer} compares Token, Alignment, and GL-JEPA under a shared protocol without Graph Context Pretraining.
All settings are shared across methods, while the language-side modules remain specific to each supervision interface. A Linear Predictor variant of GL-JEPA serves as a capacity control, with nearly the same number of trainable parameters as Alignment. Frozen-transfer evaluation covers node classification and link prediction on seven datasets each, molecular graph classification on four datasets, and full-pool graph--text retrieval on four datasets. Appendix~\ref{app:rq1-protocol} provides the complete training and evaluation protocols.

\begin{wrapfigure}{r}{0.32\textwidth}
  \vspace{-5mm}
  \centering
  \captionof{table}{Controlled comparison of graph--language supervision interfaces (\%). Trainable parameters include the Graph Encoder.}
  \vspace{-3mm}
  \label{tab:frozen-transfer}
  \scriptsize
  \setlength{\tabcolsep}{2pt}
  \renewcommand{\arraystretch}{1.2}
  \resizebox{\linewidth}{!}{%
    \begin{tabular}{lccccc}
      \toprule
      Method
      & \shortstack{Trainable\\Params.(M)}
      & \shortstack{Node\\(Acc.)}
      & \shortstack{Link\\(AUC)}
      & \shortstack{Retrieval\\(R@1)}
      & \shortstack{Graph\\(AUC)} \\
      \midrule

      Token
      & 1,031
      & \textcolor{lowergray}{25.62}
      & \textcolor{lowergray}{67.37}
      & \textcolor{lowergray}{42.52}
      & \textcolor{lowergray}{39.45} \\

      Alignment
      & 18.5
      & \textcolor{lowergray}{\textcolor{black}{55.96}}
      & \textcolor{lowergray}{\textcolor{black}{71.90}}
      & \textbf{\textcolor{black}{75.07}}
      & \textcolor{lowergray}{\textcolor{black}{51.78}} \\

      \rowcolor{gljepablue}
      \textbf{GL-JEPA}
      & 772.6
      & \textbf{\textcolor{black}{59.55}}
      & \textbf{\textcolor{black}{75.72}}
      & \textcolor{black}{67.36}
      & \textbf{\textcolor{black}{54.33}} \\

      \midrule
      \textcolor{lowergray}{\shortstack{GL-JEPA\\(Linear Pred.)}}
      & 18.7
      & \textcolor{lowergray}{\textcolor{black}{58.40}}
      & \textcolor{lowergray}{\textcolor{black}{73.94}}
      & \textcolor{lowergray}{\textcolor{black}{65.32}}
      & \textcolor{lowergray}{\textcolor{black}{53.02}} \\

      \bottomrule
    \end{tabular}%
  }
   \vspace{-3mm}
\end{wrapfigure}

   \vspace{-2mm}
\paragraph{Results.}
Token performs worst across all four transfer metrics, \textcolor{black}{consistent with its objective of reconstructing a particular linguistic realization rather than directly optimizing graph-task representations}. Alignment achieves the highest Retrieval R@1, as it directly optimizes graph--text matching, but trails GL-JEPA on Node, Link, and Graph. \textcolor{black}{This pattern is consistent with matching providing weaker supervision for graph features that are not needed for retrieval.} The nearly parameter-matched Linear Predictor shows the same advantage over Alignment on graph tasks, indicating that the pattern is not explained by Predictor parameter count alone.

\subsection{Continuous Supervision for Language Prediction}
\label{sec:continuous-supervision}

\paragraph{Setup.}
Figure~\ref{fig:dpa-efficiency} evaluates language prediction and training cost under matched pretraining data and sample counts.
Token and GL-JEPA use the same paired graph--text pretraining data.
\textcolor{black}{For METEOR, a separately trained generation adapter connects the frozen GL-JEPA checkpoint to a frozen decoder. This readout is not part of pretraining.}
Language prediction is evaluated on WebNLG with METEOR at matched numbers of processed pretraining samples.
Dual-positive accuracy (DPA) measures whether two valid descriptions both outrank the same negative on WebNLG and DART \textcolor{black}{under each model's native scoring rule}.
Training and inference cost are measured in time per sample. Appendix~\ref{app:text-readout} provides the scoring, generation, and timing protocols.

\paragraph{Results.}
Because token prediction is trained against a single observed wording, it directly rewards reproducing that linguistic realization. GL-JEPA instead predicts a continuous description representation, allowing semantically valid alternatives to be recognized without matching the reference token sequence. GL-JEPA achieves higher WebNLG METEOR at matched pretraining sample counts and improves dual-positive accuracy by 25.23 points on WebNLG and 43.75 points on DART. It also reduces measured training-update cost by 66.0\%, including generation-readout fitting, by avoiding token-level reconstruction during pretraining.

\subsection{Feature Suppression Across Supervision Objectives}
\label{sec:feature-suppression}

\paragraph{Setup.}
We test whether Alignment preserves information that is present in the target but unnecessary for distinguishing easy graph--text pairs \citep{robinson2021shortcut,li2023featuresuppression}. We use two diagnostics. In the synthetic setting, each graph contains two factors, \(A\) and \(B\), encoded in node features. Pairs such as \((A_1,B_3)\) and \((A_7,B_5)\) can be distinguished using \(A\) alone, whereas \((A_1,B_3)\) and \((A_1,B_5)\) require \(B\). In WikiBio, we construct two-node graphs from nationality and occupation. Easy candidates differ in nationality, while hard candidates share nationality and differ only in occupation. We compare easy-negative Alignment, hard-negative Alignment, and GL-JEPA latent prediction. After pretraining, each Graph Encoder is frozen and a linear probe measures how well \(B\) or occupation can be recovered from its graph states. Appendix~\ref{app:feature-suppression} gives the full protocols.

\begin{wrapfigure}{r}{0.32\textwidth}
  \centering
  \vspace{-4mm}
  \includegraphics[width=\linewidth]{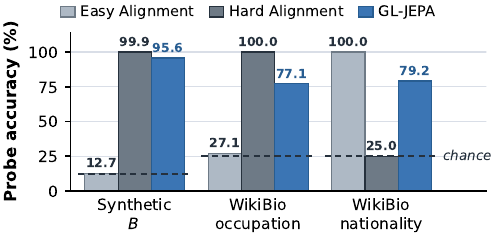}
  \vspace{-6mm}
  \caption{Linear-probe accuracy for \(B\) and occupation. GL-JEPA denotes latent prediction. Dashed lines mark chance.}
  \label{fig:feature-suppression}
  \vspace{-5mm}
\end{wrapfigure}
\paragraph{Results.}
With easy negatives, the model can identify the matching description using only \(A\) in the synthetic setting or nationality in WikiBio. As a result, \(B\) and occupation contribute little to the matching objective and are nearly unrecoverable from the frozen graph states. When the easy cue is held fixed, the model must use \(B\) or occupation to distinguish the correct match, and probe accuracy rises sharply with the same encoder architecture. GL-JEPA does not require such factor-specific negatives: by predicting the full target, it retains substantially more information about \(B\) and occupation than easy-negative Alignment, although hard-negative Alignment remains higher.

\subsection{Frozen Transfer Across Graph Tasks}
\label{sec:zsl-retrieval}

\paragraph{Setup.}
Table~\ref{tab:all-evaluation} compares GL-JEPA with GraphCLIP \citep{zhu2025graphclip}, ADAligner \citep{liu2026adaligner}, and G2P2 \citep{wen2023g2p2} under a shared pretraining corpus and frozen-transfer evaluation protocol. All methods are retrained on the same graph--text data while retaining their original architectures and training objectives. Evaluation covers node classification and link prediction on seven datasets (Cora, CiteSeer, WikiCS, Instagram, Photo, Computers, and History), graph--text retrieval on four datasets (Agenda, GenWiki, TekGen, and T-REx), and graph classification on four molecular datasets (ClinTox, HIV, Tox21, and ToxCast). All tasks follow a shared CLIP-style protocol based on embedding compatibility with class descriptions, edge endpoints, or graph descriptions, without updating model parameters on the evaluation datasets. We report mean accuracy for Node, mean AUC for Link and Graph, and bidirectional R@1 for Retrieval. Appendix~\ref{app:shared-evaluation-contract} provides the full pretraining and evaluation protocols.

\begin{table}[t]
  \caption{Frozen transfer across graph tasks (\%). Best values are bold. GL-JEPA is shaded blue. Values lower than GL-JEPA are gray.}
  \vspace{-3mm}
  \label{tab:all-evaluation}
  \centering
  \scriptsize
  \setlength{\tabcolsep}{4.5pt}
  \renewcommand{\arraystretch}{1.03}

  \resizebox{\textwidth}{!}{%
    \begin{tabular}{lccccccccccccc}
      \toprule
      \textbf{Model}
      & \multicolumn{8}{c}{\textbf{Node Classification Acc. (\%)}}
      & \multicolumn{5}{c}{\textbf{Bidirectional Retrieval R@1 (\%)}} \\
      \cmidrule(lr){2-9}
      \cmidrule(l){10-14}

      & \textbf{Cora}
      & \textbf{Cite.}
      & \textbf{Wiki}
      & \textbf{Insta.}
      & \textbf{Photo}
      & \textbf{Comp.}
      & \textbf{Hist.}
      & \textbf{Mean}
      & \textbf{Agenda}
      & \textbf{GenW.}
      & \textbf{TekGen}
      & \textbf{T-REx}
      & \textbf{Mean} \\
      \midrule

      GraphCLIP
      & \textcolor{lowergray}{62.94}
      & \textcolor{lowergray}{61.07}
      & \textbf{70.43}
      & \textcolor{lowergray}{50.46}
      & \textcolor{lowergray}{52.38}
      & \textbf{60.35}
      & \textcolor{lowergray}{38.92}
      & \textcolor{lowergray}{56.65}
      & 73.89
      & \textcolor{lowergray}{81.33}
      & \textbf{91.07}
      & \textcolor{lowergray}{55.91}
      & \textbf{75.55} \\

      ADAligner
      & \textcolor{lowergray}{57.06}
      & \textcolor{lowergray}{59.03}
      & \textcolor{lowergray}{63.27}
      & \textcolor{lowergray}{51.04}
      & \textcolor{lowergray}{54.92}
      & \textcolor{lowergray}{40.68}
      & \textcolor{lowergray}{38.05}
      & \textcolor{lowergray}{52.01}
      & \textbf{75.30}
      & \textcolor{lowergray}{74.21}
      & \textcolor{lowergray}{81.45}
      & \textcolor{lowergray}{51.29}
      & \textcolor{lowergray}{70.56} \\

      G2P2
      & \textcolor{lowergray}{58.05}
      & \textcolor{lowergray}{59.17}
      & \textcolor{lowergray}{34.03}
      & \textbf{59.36}
      & \textcolor{lowergray}{25.48}
      & \textcolor{lowergray}{37.42}
      & \textcolor{lowergray}{26.27}
      & \textcolor{lowergray}{42.83}
      & \textcolor{lowergray}{66.05}
      & \textcolor{lowergray}{45.11}
      & \textcolor{lowergray}{66.83}
      & \textbf{60.79}
      & \textcolor{lowergray}{59.70} \\

      \rowcolor{gljepablue}\textbf{\method{}}
      & \textbf{67.36}
      & \textbf{67.43}
      & 68.27
      & 57.65
      & \textbf{56.44}
      & 59.48
      & \textbf{48.26}
      & \textbf{60.70}
      & 71.28
      & \textbf{81.34}
      & 83.16
      & 57.87
      & 73.41 \\

      \bottomrule
    \end{tabular}%
  }

  \vspace{-0.2em}

  \resizebox{\textwidth}{!}{%
    \begin{tabular}{lccccccccccccc}
      \toprule
      \textbf{Model}
      & \multicolumn{8}{c}{\textbf{Link Prediction AUC (\%)}}
      & \multicolumn{5}{c}{\textbf{Graph Classification AUC (\%)}} \\
      \cmidrule(lr){2-9}
      \cmidrule(l){10-14}

      & \textbf{Cora}
      & \textbf{Cite.}
      & \textbf{Wiki}
      & \textbf{Insta.}
      & \textbf{Photo}
      & \textbf{Comp.}
      & \textbf{Hist.}
      & \textbf{Mean}
      & \textbf{ClinTox}
      & \textbf{HIV}
      & \textbf{Tox21}
      & \textbf{ToxCast}
      & \textbf{Mean} \\
      \midrule

      GraphCLIP
      & \textcolor{lowergray}{72.36}
      & \textcolor{lowergray}{70.95}
      & \textcolor{lowergray}{84.86}
      & \textbf{67.20}
      & \textcolor{lowergray}{60.28}
      & \textcolor{lowergray}{66.95}
      & \textcolor{lowergray}{78.53}
      & \textcolor{lowergray}{71.59}
      & \textcolor{lowergray}{44.46}
      & \textcolor{lowergray}{54.81}
      & \textcolor{lowergray}{63.28}
      & \textbf{61.55}
      & \textcolor{lowergray}{56.03} \\

      ADAligner
      & \textcolor{lowergray}{55.88}
      & \textcolor{lowergray}{55.55}
      & \textcolor{lowergray}{55.20}
      & \textcolor{lowergray}{50.97}
      & \textcolor{lowergray}{51.09}
      & \textcolor{lowergray}{51.21}
      & \textcolor{lowergray}{53.15}
      & \textcolor{lowergray}{53.29}
      & \textcolor{lowergray}{46.72}
      & \textcolor{lowergray}{54.49}
      & \textcolor{lowergray}{60.86}
      & \textcolor{lowergray}{54.45}
      & \textcolor{lowergray}{54.13} \\

      G2P2
      & \textcolor{lowergray}{74.99}
      & \textcolor{lowergray}{71.77}
      & \textcolor{lowergray}{80.25}
      & 63.57
      & \textcolor{lowergray}{71.50}
      & \textcolor{lowergray}{68.41}
      & \textcolor{lowergray}{80.92}
      & \textcolor{lowergray}{73.06}
      & \textbf{51.66}
      & \textcolor{lowergray}{39.55}
      & \textcolor{lowergray}{43.29}
      & \textcolor{lowergray}{44.86}
      & \textcolor{lowergray}{44.84} \\

      \rowcolor{gljepablue}\textbf{\method{}}
      & \textbf{77.35}
      & \textbf{75.05}
      & \textbf{85.18}
      & 57.99
      & \textbf{73.90}
      & \textbf{76.97}
      & \textbf{82.63}
      & \textbf{75.58}
      & 50.60
      & \textbf{57.19}
      & \textbf{65.25}
      & 59.55
      & \textbf{58.15} \\

      \bottomrule
    \end{tabular}%
  }  \vspace{-3mm}

\end{table}

\paragraph{Results.}
GL-JEPA achieves the highest mean scores on Node, Link, and Graph tasks, outperforming the strongest baselines by 4.05 percentage points in Node accuracy and 2.52 percentage points in Link AUC, while attaining a mean Graph AUC of 58.15\%. Its advantage is particularly consistent on Link, where it leads on six of seven datasets, although the strongest method varies across individual Node and Graph datasets. In contrast, GraphCLIP achieves higher mean Retrieval R@1 (75.55\% versus 73.41\% for GL-JEPA), with dataset-level performance varying across methods.

\subsection{Graph Question Answering}
\label{sec:graph-question-answering}

\paragraph{Setup.}
Table~\ref{tab:graphqa-five-task} evaluates graph question answering (GQA) following latent predictive pretraining without autoregressive text reconstruction. We compare GL-JEPA with GraphGPT \citep{tang2023graphgpt}, LLaGA \citep{chen2024llaga}, GraphWiz \citep{chen2024graphwiz}, GraphSolver \citep{luo2024graphinstruct}, GOFA \citep{kong2025gofa}, TEA-GLM \citep{chen2024teaglm}, and G1 \citep{guo2025g1}.
During GQA adaptation, GL-JEPA keeps the pretrained Graph Encoder \(E_G\) frozen while adapting the Predictor \(P_\omega\) and its output projection.
The Predictor conditions on the frozen graph representations and task query to score candidate answers, following the discriminative visual question answering protocol of VL-JEPA \citep{chen2025vljepa}, rather than generating free-form text. Evaluation covers five GraphInstruct tasks—Bipartite, Connectivity, Cycle, Hamiltonian, and Substructure—with 400 test examples each, as well as ExplaGraphs, CLeGR-Facts, and the Connectivity and Cycle subsets of NLGraph. Appendix~\ref{app:query-conditioned-qa} provides the adaptation and scoring protocols.

\paragraph{Results.}
\textcolor{black}{GL-JEPA achieves 79.75\% mean accuracy across the five GraphInstruct tasks with \(E_G\) frozen, compared with the 77.40\% reported in the GraphWiz paper. Among the results, GL-JEPA achieves the highest accuracy on Hamiltonian and Substructure and exceeds the strongest listed baselines on NLGraph Connectivity and Cycle by 13.20 and 39.27 percentage points, respectively. Other systems report higher scores on GraphInstruct Bipartite, Connectivity, and Cycle, as well as ExplaGraphs and CLeGR-Facts. \textcolor{black}{These results demonstrate that GL-JEPA supports competitive query-conditioned graph question answering without updating the pretrained Graph Encoder.}}

\begin{table}[t]
  \caption{Graph question answering accuracy (\%) across benchmarks. GraphInstruct Mean averages the five GraphWiz tasks. GL-JEPA is shaded light blue. Values lower than GL-JEPA are gray.}
  \label{tab:graphqa-five-task}
  \vspace{-3mm}

  \centering
  \scriptsize
  \setlength{\tabcolsep}{4.7pt}
  \renewcommand{\arraystretch}{1.03}

  \resizebox{0.99\linewidth}{!}{%
    \begin{tabular}{lccccccccccc}
      \toprule

      \multirow{2}{*}{\textbf{Model}}
      &
      & \multicolumn{6}{c}{\textbf{GraphInstruct}}
      & \multirow{2}{*}{\textbf{Expla.}}
      & \textbf{CLeGR}
      & \multicolumn{2}{c}{\textbf{NLGraph}} \\

      \cmidrule(lr){3-8}
      \cmidrule(l){10-10}
      \cmidrule(lr){11-12}

      & \textbf{Backbone Size}
      & \textbf{Bip.}
      & \textbf{Conn.}
      & \textbf{Cyc.}
      & \textbf{Ham.}
      & \textbf{Sub.}
      & \textbf{Mean}
      &
      & \textbf{Facts}
      & \textbf{Conn.}
      & \textbf{Cyc.} \\

      \midrule

      GraphGPT
      & 7B
      & \textbf{92.75}
      & \textcolor{lowergray}{70.75}
      & \textcolor{lowergray}{71.25}
      & \textcolor{lowergray}{60.50}
      & \textcolor{lowergray}{80.50}
      & \textcolor{lowergray}{75.15}
      & \textcolor{lowergray}{55.05}
      & \textcolor{lowergray}{85.61}
      & \textcolor{lowergray}{49.87}
      & \textcolor{lowergray}{54.97} \\

      LLaGA
      & 7B
      & \textcolor{lowergray}{71.25}
      & \textcolor{lowergray}{68.00}
      & \textcolor{lowergray}{68.75}
      & \textcolor{lowergray}{58.75}
      & \textcolor{lowergray}{45.25}
      & \textcolor{lowergray}{62.40}
      & \textcolor{lowergray}{51.26}
      & \textcolor{lowergray}{88.16}
      & \textcolor{lowergray}{50.13}
      & \textcolor{lowergray}{51.31} \\

      GraphWiz
      & 7B
      & 74.00
      & \textbf{87.00}
      & 91.50
      & \textcolor{lowergray}{52.25}
      & \textcolor{lowergray}{82.25}
      & \textcolor{lowergray}{77.40}
      & \textbf{77.98}
      & \textcolor{lowergray}{84.76}
      & \textcolor{lowergray}{70.89}
      & \textcolor{lowergray}{60.21} \\

      GraphSolver
      & 8B
      & \textcolor{lowergray}{64.50}
      & \textcolor{lowergray}{67.75}
      & \textcolor{lowergray}{68.50}
      & \textcolor{lowergray}{31.75}
      & \textcolor{lowergray}{41.75}
      & \textcolor{lowergray}{54.85}
      & \textcolor{lowergray}{58.66}
      & \textbf{95.61}
      & \textcolor{lowergray}{64.15}
      & \textcolor{lowergray}{52.88} \\

      GOFA
      & 7B
      & \textcolor{lowergray}{68.00}
      & \textcolor{lowergray}{67.75}
      & \textcolor{lowergray}{68.75}
      & \textcolor{lowergray}{32.25}
      & \textcolor{lowergray}{47.50}
      & \textcolor{lowergray}{56.85}
      & \textcolor{lowergray}{58.84}
      & \textcolor{lowergray}{79.21}
      & \textcolor{lowergray}{53.64}
      & \textcolor{lowergray}{52.88} \\

      TEA-GLM
      & 7B
      & \textcolor{lowergray}{33.25}
      & \textcolor{lowergray}{61.75}
      & \textcolor{lowergray}{55.50}
      & \textcolor{lowergray}{28.75}
      & \textcolor{lowergray}{59.75}
      & \textcolor{lowergray}{47.80}
      & \textcolor{lowergray}{59.57}
      & \textcolor{lowergray}{88.07}
      & \textcolor{lowergray}{40.70}
      & \textcolor{lowergray}{36.13} \\

      G1-3B
      & 3B
      & \textcolor{lowergray}{64.50}
      & \textcolor{lowergray}{40.25}
      & \textbf{93.50}
      & \textcolor{lowergray}{49.25}
      & \textcolor{lowergray}{40.25}
      & \textcolor{lowergray}{57.55}
      & \textcolor{lowergray}{44.22}
      & \textcolor{lowergray}{79.66}
      & \textcolor{lowergray}{73.05}
      & \textcolor{lowergray}{52.88} \\

      \addlinespace[0.15em]

      \rowcolor{gljepablue}
      \textbf{\method{}}
      & 1B
      & 73.75
      & 86.75
      & 89.75
      & \textbf{62.75}
      & \textbf{85.75}
      & \textbf{79.75}
      & 61.01
      & 93.58
      & \textbf{86.25}
      & \textbf{99.48} \\

      \bottomrule
    \end{tabular}%
  }

  \vspace{-3mm}
\end{table}

\begin{table}[H]
  \centering
  \caption{Ablation studies of GL-JEPA (\%). GQA averages the five GraphInstruct tasks.}

  \label{tab:ablation}
  \scriptsize
  \setlength{\tabcolsep}{2.5pt}
  \renewcommand{\arraystretch}{1.03}
  \begin{minipage}[t]{0.49\linewidth}
    \begin{tabular*}{\linewidth}{@{\extracolsep{\fill}}lcccc@{}}
      \toprule
      Configuration & Node & Link & Retrieval & GQA \\
      % section divider removed
      \midrule
      \multicolumn{5}{l}{\textit{(a) Pretraining components}} \\
      % divider removed
      \rowcolor{gljepablue}[2pt][2pt]\textbf{Full pipeline} & 60.70 & 75.58 & 73.41 & 79.75 \\
      w/o Graph Context Pretraining & \textcolor{lowergray}{58.90} & \textcolor{lowergray}{75.22} & \textcolor{lowergray}{68.72} & \textcolor{lowergray}{77.75} \\
      Graph Context Pretraining only & \textcolor{lowergray}{43.14} & \textcolor{lowergray}{63.42} & \textcolor{lowergray}{2.06} & \textcolor{lowergray}{61.70} \\
      w/o Pretraining & \textcolor{lowergray}{45.60} & \textcolor{lowergray}{71.62} & \textcolor{lowergray}{2.52} & \textcolor{lowergray}{62.30} \\
      % divider removed
      \midrule
      \multicolumn{5}{l}{\textit{(b) Prediction objective}} \\
      % divider removed
      \rowcolor{gljepablue}[2pt][2pt]\textbf{MSE} & 60.70 & 75.58 & 73.41 & 79.75 \\
      Smooth-L1 & 60.92 & \textcolor{lowergray}{65.10} & \textcolor{lowergray}{64.03} & \textcolor{lowergray}{76.35} \\
      InfoNCE & \textcolor{lowergray}{60.42} & \textcolor{lowergray}{63.89} & 78.36 & \textcolor{lowergray}{78.00} \\
      % divider removed
      \midrule
      \multicolumn{5}{l}{\textit{(c) Text Encoder variation}} \\
      % divider removed
      \rowcolor{gljepablue}[2pt][2pt]\textbf{all-MiniLM-L6-v2 (default)} & 60.70 & 75.58 & 73.41 & 79.75 \\
      EmbeddingGemma-300M & 62.82 & \textcolor{lowergray}{71.40} & 74.14 & \textcolor{lowergray}{74.75} \\
      Qwen3-Embedding-0.6B & 61.25 & \textcolor{lowergray}{75.17} & \textcolor{lowergray}{70.81} & 79.80 \\
      BGE-M3 (568M) & 62.06 & \textcolor{lowergray}{73.78} & \textcolor{lowergray}{69.63} & \textcolor{lowergray}{75.05} \\
      \bottomrule
    \end{tabular*}
  \end{minipage}\hfill
  \begin{minipage}[t]{0.49\linewidth}
    \begin{tabular*}{\linewidth}{@{\extracolsep{\fill}}lcccc@{}}
      \toprule
      Configuration & Node & Link & Retrieval & GQA \\
      % divider removed
      \midrule
      \multicolumn{5}{l}{\textit{(d) Predictor architecture}} \\
      % section rule removed
      \rowcolor{gljepablue}[2pt][2pt]\textbf{Llama-initialized Transformer} & 60.70 & 75.58 & 73.41 & 79.75 \\
      Random Transformer & \textcolor{lowergray}{60.22} & \textcolor{lowergray}{63.88} & \textcolor{lowergray}{72.02} & \textcolor{lowergray}{78.00} \\
      MLP & \textcolor{lowergray}{56.86} & \textcolor{lowergray}{71.02} & \textcolor{lowergray}{57.91} & \textcolor{lowergray}{71.50} \\
      Linear & \textcolor{lowergray}{59.07} & \textcolor{lowergray}{72.19} & \textcolor{lowergray}{63.51} & \textcolor{lowergray}{74.10} \\
      Attention + MLP & \textcolor{lowergray}{57.07} & \textcolor{lowergray}{63.14} & \textcolor{lowergray}{62.10} & \textcolor{lowergray}{71.05} \\
      % section rule removed
      \midrule
      \multicolumn{5}{l}{\textit{(e) Predictor input tokens}} \\
      % section rule removed
      \rowcolor{gljepablue}[2pt][2pt]\textbf{Graph + node tokens} & 60.70 & 75.58 & 73.41 & 79.75 \\
      Graph token only & \textcolor{lowergray}{59.75} & \textcolor{lowergray}{71.26} & \textcolor{lowergray}{55.78} & \textcolor{lowergray}{72.30} \\
      Node tokens only & \textcolor{lowergray}{60.22} & \textcolor{lowergray}{68.04} & \textcolor{lowergray}{62.14} & 80.45 \\
      % section rule removed
      \midrule
      \multicolumn{5}{l}{\textit{(f) Graph Encoder variation}} \\
      % section rule removed
      \rowcolor{gljepablue}[2pt][2pt]\textbf{GraphGPS (default)} & 60.70 & 75.58 & 73.41 & 79.75 \\
      GraphSAGE (4 layers) & 61.12 & 77.01 & \textcolor{lowergray}{66.63} & \textcolor{lowergray}{79.25} \\
      GCN (4 layers) & 60.94 & 77.32 & \textcolor{lowergray}{68.25} & \textcolor{lowergray}{78.65} \\
      \bottomrule
    \end{tabular*}
  \end{minipage}
\end{table}

\subsection{Ablation of GL-JEPA Design Choices}
\label{sec:ablation}

\paragraph{Setup.}
We conduct six ablation studies to examine how GL-JEPA's pretraining stages, prediction objective, Text Encoder, Predictor architecture, Predictor input tokens, and Graph Encoder affect transfer across graph prediction, graph--text retrieval, and graph question answering (Table~\ref{tab:ablation}). Within each ablation group, we hold all settings except for changes necessitated by removing a pretraining stage. Node accuracy, Link AUC, and Retrieval R@1 are evaluated with the Graph Encoder frozen, while GQA reports mean accuracy after adapting the Predictor with the Graph Encoder frozen. Appendices~\ref{app:shared-evaluation-contract} and~\ref{app:implementation-details} detail the evaluation and training protocols.

\textbf{(a) Pretraining components.}
The full pipeline outperforms the variant without Graph Context Pretraining across all four metrics, showing the benefit of structural pretraining before description-target prediction. 

\textbf{(b) Prediction objective.}
InfoNCE improves Retrieval but not graph prediction or question answering. Node accuracy changes little, while Link AUC drops substantially. 

\textbf{(c) Text Encoder variation.}
Different Text Encoders yield task-dependent results, showing that the continuous description target affects transfer beyond encoder size. 

\textbf{(d) Predictor architecture.}
Random initialization mainly affects Link performance, whereas replacing the Transformer degrades transfer across more tasks. Predictor architecture contributes beyond initialization. 

\textbf{(e) Predictor input tokens.}
Graph-summary and graph-node tokens are complementary for frozen transfer. After query-conditioned adaptation, node tokens alone are sufficient for GQA, and additional tokens do not improve performance. 

\textbf{(f) Graph Encoder variation.}
GraphSAGE and GCN variants \citep{kipf2017semi} perform well on local graph prediction but worse on graph--text retrieval, showing that Graph Encoder architecture affects transfer under the same continuous description target.

\subsection{Case Study}
\label{sec:case-study}

\begin{wrapfigure}{r}{0.6\textwidth}
  \centering
  \vspace{-6mm}
  \includegraphics[width=\linewidth]{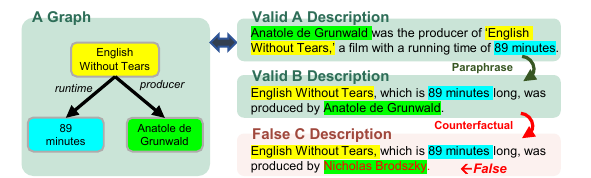}
    \vspace{-3mm}

  \caption{Case study of dual-positive accuracy on a held-out WebNLG example.}
  \label{fig:case-study}
  \vspace{-4mm}
\end{wrapfigure}

\paragraph{Interpretation.}
Figure~\ref{fig:case-study} shows a held-out graph encoding that Anatole de Grunwald produced \emph{English Without Tears} and that the film runs for 89 minutes. A and B express these same facts with different wording, while C changes the producer in B's sentence. Token ranks them \(A_{\mathrm{valid}} > C_{\mathrm{false}} > B_{\mathrm{valid}}\), placing the false description above the valid paraphrase B. GL-JEPA ranks them \(A_{\mathrm{valid}} > B_{\mathrm{valid}} > C_{\mathrm{false}}\), placing both valid descriptions above the false one. GL-JEPA preserves the factual ordering across paraphrases, while Token does not in this case.

\section{Related Work}
\label{sec:related}

\paragraph{Graph--Language Models.}

Autoregressive graph--language methods condition language decoders on graph representations and optimize token-level generation losses \citep{tang2023graphgpt,chen2024llaga}. Graph context is incorporated through learned tokens, serialized neighborhoods, or retrieved subgraphs \citep{he2024gretriever}. Approaches employ instruction tuning for graph question answering and multi-task understanding \citep{chen2024graphwiz,luo2024graphinstruct}, extend generation across graph task levels \citep{kong2025gofa}, align GNN representations with language-model token embeddings \citep{chen2024teaglm}, or use reinforcement learning for graph reasoning \citep{guo2025g1}. GL-JEPA predicts a continuous description representation during pretraining and trains a decoder only for generation.

\paragraph{Graph--Text Alignment.}
Graph--text alignment learns compatibility between matched and mismatched graph--description pairs \citep{wen2023g2p2,wang2023congrat}, enabling the transfer of textual supervision to graph representations across datasets and tasks \citep{zhu2025graphclip,fang2024uniglm,liu2026adaligner,wang2026s2aligner}. A contrastive objective can be satisfied by features sufficient for pair discrimination even when other target-relevant features remain weakly represented \citep{robinson2021shortcut,li2023featuresuppression}. Prior work has explored latent target decoding to mitigate predictive feature suppression in image--caption retrieval \citep{bleeker2022predictive}. GL-JEPA predicts a frozen description representation from graph context rather than directly aligning pooled graph and text embeddings.

\paragraph{Joint-Embedding Predictive Architectures.}
Joint-embedding predictive methods estimate target representations from observed context rather than reconstructing inputs
\citep{grill2020byol,baevski2022data2vec}.
Image methods predict representations of missing regions without hand-designed augmentation
\citep{assran2023self},
while video methods extend latent prediction across time for understanding and planning
\citep{bardes2024vjepa,assran2025vjepa2}.
\textcolor{black}{Related objectives have been applied to audio, point clouds, and tables
\citep{fei2023ajepa,saito2025pointjepa,thimonier2025tjepa}, as well as language
\citep{huang2025llmjepa} and multimodal alignment \citep{lei2025m3jepa}.}
\textcolor{black}{Graph-level work predicts masked-subgraph representations \citep{skenderi2025graphjepa}, while vision--language work predicts text embeddings from visual context \citep{chen2025vljepa}. GL-JEPA predicts a frozen description target from graph structure and node text. We compare its graph-task transfer with autoregressive token prediction and direct graph--text alignment.}

\section{Discussion}

Compared with autoregressive token prediction, GL-JEPA replaces token-level reconstruction with prediction of a continuous description target. Token is trained to reproduce the particular wording of the paired description, whereas latent prediction targets its frozen representation. In the controlled comparison and case study, latent prediction more often ranks differently worded valid descriptions above an incorrect one. These results suggest less dependence on the wording of the training description.

Compared with direct graph--text Alignment, latent prediction targets the full description representation rather than only distinguishing matched from mismatched pairs. In the synthetic and WikiBio diagnostics, target-relevant factors that matching need not use are more recoverable from the frozen graph states of the latent-prediction model than from those of Alignment, indicating less feature suppression in these controlled settings. Separately, Alignment's higher Retrieval R@1 does not translate into better Node, Link, or Graph transfer in the controlled comparison.

These findings point to why predicting a frozen description target is useful. The Graph Encoder and Predictor learn to estimate the Text Encoder's representation of the paired description, so pretraining does not reconstruct the observed wording and can supervise factors encoded in that representation even when pair matching does not require them.

\textbf{Limitations and Future Work.}
GL-JEPA predicts a single frozen description representation per training example, limiting language supervision to the information encoded in that target. Future work could predict multiple description representations to capture more semantic information. Multi-step latent prediction could combine intermediate Predictor outputs with graph states to refine subsequent predictions before task-specific readout.

\section{Conclusion}

We introduced GL-JEPA, a graph--language pretraining framework that predicts frozen description representations from graph states without autoregressive token reconstruction or direct graph--text alignment. Compared with Token, GL-JEPA ranks two valid descriptions above a negative without reconstructing their wording. In controlled comparisons, GL-JEPA outperforms Alignment on Node, Link, and Graph tasks, whereas Alignment achieves higher Retrieval R@1. In the controlled feature-suppression diagnostic, a target-relevant factor is more recoverable from latent-prediction graph states than from Alignment graph states when matching does not require that factor. With the Graph Encoder frozen, the learned representations also support text generation through a separate readout and query-conditioned graph question answering through Predictor adaptation.

\section*{AI Use Statement}

We used generative AI tools, including OpenAI Codex, to assist with experimental-design feedback, implementation and debugging of model and analysis code, construction of figures and tables, and interpretation of results. These tools also assisted in implementing the procedural generator for the synthetic feature-suppression diagnostic, which samples graph factors \(A\), \(B\), and \(S\) under the protocol in Appendix~\ref{app:feature-suppression}. They further assisted with literature discovery, citation checks, translation, and drafting and editing parts of the manuscript. The authors decided which suggestions and experiments to adopt, reviewed the AI-assisted outputs, and take responsibility for the final methods, results, claims, and text.

\section*{Ethics Statement}

This work uses previously released graph datasets, graph--description pairs, and synthetic graphs for representation learning and evaluation.
We do not collect new human-subject data.
Some pretraining data, including social-network graphs, may contain user-generated text and may inherit biases or other artifacts from the original sources and their description-generation pipelines.
GL-JEPA may learn or propagate such biases through its graph and language representations.
The experiments focus on representation learning and benchmark evaluation rather than deployment in high-stakes settings.
We do not use the learned representations to infer sensitive attributes or to identify individuals.

\section*{Reproducibility Statement}

The manuscript specifies the data, models, and procedures used for every
reported evaluation. Appendix~\ref{app:datasets-eval-coverage} identifies the
five graph--description corpora, their graph counts, and the provenance of the
released descriptions. Appendix~\ref{app:shared-evaluation-contract} fixes the
downstream datasets, stored splits, retrieval candidate pools, seeds, and
summary metrics. Appendix~\ref{app:feature-suppression} gives the complete
construction and probe protocol for the synthetic diagnostic. The model
configuration, two-stage pretraining schedule, optimization settings, and
ablation controls are specified in Appendix~\ref{app:implementation-details},
while Appendix~\ref{app:rq1-protocol} documents the task-specific readouts,
adaptation, and scoring procedures. Training and evaluation code, data
manifests, and checkpoint identifiers will accompany the released artifact.

\bibliography{references}
\bibliographystyle{iclr2027_conference}

\clearpage
\appendix
\clearpage
\section{Datasets and Evaluation Coverage}
\label{app:datasets-eval-coverage}

\begin{table}[H]
\centering
\caption{Graph counts in the GraphCLIP pretraining corpora used by GL-JEPA.}
\label{tab:pretraining-statistics}
\scriptsize
\begin{tabular}{lr}
\toprule
Corpus & Graphs \\
\midrule
PubMed & 19,717 \\
OGBN-Arxiv & 169,343 \\
ArXiv-2023 & 46,198 \\
Reddit & 33,434 \\
OGBN-Products & 196,615 \\
\bottomrule
\end{tabular}
\end{table}

The graph--description pairs are drawn from the GraphCLIP collection \citep{zhu2025graphclip}. We use its released target strings verbatim, including the Qwen-based generation and filtering pipeline. Table~\ref{tab:pretraining-statistics} reports the number of graphs in each pretraining corpus. Downstream datasets, task families, and metrics are summarized in Appendix~\ref{app:shared-evaluation-contract}.

The corpora span three graph domains. PubMed, OGBN-Arxiv, and ArXiv-2023 provide scientific citation graphs. Reddit provides social interaction graphs, and OGBN-Products provides product co-purchase graphs. This mixture exposes the Graph Encoder to different node vocabularies, neighborhood structures, and relation semantics before downstream evaluation. Each stored training example consists of a text-attributed graph and its released paired description. Node text is encoded as graph input, whereas the paired description is used only as the target during Graph--Language Latent Prediction.

The counts in Table~\ref{tab:pretraining-statistics} describe the graph examples available to pretraining, rather than the sizes of the original source networks. Graph sampling retains directed edge information. \textcolor{black}{The controlled Token, Alignment, and GL-JEPA comparison uses the same materialized graph--description examples and varies the language-supervision interface.}

\section{Shared Evaluation Contract}
\label{app:shared-evaluation-contract}

\begin{table}[H]
\centering
\caption{Task families and summary metrics used in the main evaluation.}
\label{tab:evaluation-summary}
\small
\begin{tabular}{p{0.22\linewidth}p{0.48\linewidth}p{0.20\linewidth}}
\toprule
Task family & Datasets or benchmarks & Reported metric \\
\midrule
Node classification & Cora, CiteSeer, WikiCS, Instagram, Photo, Computers, History & Mean accuracy \\
Link prediction & Cora, CiteSeer, WikiCS, Instagram, Photo, Computers, History & Mean AUC \\
Graph classification & ClinTox, HIV, Tox21, ToxCast & Mean AUC \\
Graph--text retrieval & Agenda, GenWiki, TekGen, T-REx & Bidirectional R@1 \\
Graph question answering & GraphInstruct (GraphWiz), ExplaGraphs, CLeGR-Facts, NLGraph & Accuracy \\
\bottomrule
\end{tabular}
\end{table}

Table~\ref{tab:frozen-transfer} compares supervision interfaces without Graph Context Pretraining,
Table~\ref{tab:all-evaluation} compares complete pretrained systems, and
Table~\ref{tab:ablation} evaluates GL-JEPA configurations. Retrieval uses a fixed checkpoint,
example order, and candidate pool. The \textbf{Full pipeline} row in
Table~\ref{tab:ablation} uses the Node, Link, and Retrieval means from
Table~\ref{tab:all-evaluation} and the GraphInstruct Mean from
Table~\ref{tab:graphqa-five-task}.

\textcolor{black}{For Table~\ref{tab:frozen-transfer}, each method is pretrained with seeds 90, 91, and 92. Node, Link, and Graph use five evaluation seeds (0--4) for each checkpoint. Retrieval uses fixed candidate pools. The table averages the three checkpoint scores.}

\textcolor{black}{For the complete GL-JEPA systems in Tables~\ref{tab:all-evaluation} and~\ref{tab:graphqa-five-task}, Graph Context Pretraining uses seed 88 and Graph--Language Latent Prediction uses seed 90. Table~\ref{tab:ablation} uses the same stage-specific seeds where a stage is run. Query-conditioned adaptation uses seed 90.}

\paragraph{Node classification.}
Node classification uses common cached graph views and fixed class descriptions. Accuracy is
averaged without weighting across the seven datasets.

\paragraph{Link prediction.}
Link prediction uses the same seven node datasets and five stored scoring splits. Each split uses
the same positive edges and sampled non-edges for every method. Cosine similarity between endpoint
representations supplies the link score. AUC is averaged without weighting across datasets.

\paragraph{Graph--text retrieval.}
Retrieval uses fixed full candidate pools for Agenda (1,000 examples), GenWiki (1,000), TekGen
(2,000), and T-REx (2,000). Each graph ranks every description in its dataset, and each description
ranks every graph. For dataset \(d\),
\[
 R_d=\frac{R_{G\rightarrow Y,d}+R_{Y\rightarrow G,d}}{2},
\]
and the reported mean is \(\frac14\sum_d R_d\).

\paragraph{Graph classification.}
Graph classification uses ClinTox, HIV, Tox21, and ToxCast with 80/10/10 splits for evaluation seeds 0--4 and
positive/negative class-description pairs. The score is the difference between positive and negative
class compatibility. AUC is averaged over valid binary targets containing both classes and then
without weighting across the four datasets.

\paragraph{Graph question answering.}
Graph question answering is evaluated after query-conditioned adaptation with the Graph Encoder frozen.
The GraphInstruct mean averages the accuracy of the Bipartite, Connectivity, Cycle, Hamiltonian,
and Substructure tasks introduced with GraphWiz, with 400 held-out examples per task. The adapted
checkpoint is selected using validation accuracy.

\section{Controlled Feature-Suppression Diagnostic}
\label{app:feature-suppression}

\paragraph{Data construction.}
Each generated example independently samples an easy cue $A\in\{1,\ldots,128\}$, a semantic factor $B\in\{1,\ldots,8\}$, and a structural factor $S\in\{\text{chain},\text{star}\}$.
Fixed normalized codebooks assign 16-dimensional vectors \(\mathbf c_A\) and \(\mathbf b_B\) to the two categorical factors.
Each graph has eight nodes. Node 0 receives \(5\mathbf c_A\), while nodes 1--7 receive \(0.5\mathbf b_B\).
The structural factor selects a bidirectional chain or star adjacency with self-loops.
The 138-dimensional continuous target is
\[
 \mathbf z= u\!\left([
 \operatorname{onehot}_{128}(A)\,\|\,
 \operatorname{onehot}_{8}(B)\,\|\,
 \operatorname{onehot}_{2}(S)]\right).
\]
The target contains $A$, $B$, and $S$, while the input scaling makes $A$ the easier cue.

\paragraph{Models and objectives.}
The synthetic Graph Encoder projects 16-dimensional node features to 48 dimensions, applies two message-passing layers, concatenates the mean and node-0 states, and maps the result to a two-dimensional graph state.
A two-layer MLP maps this state through a 64-dimensional hidden layer to the 138-dimensional target space.
All three conditions contain 19,580 trainable parameters.
Alignment with easy negatives applies symmetric InfoNCE with temperature 0.07 to random batches, so negatives usually differ in both $A$ and $B$. For example, $(A_1,B_3)$ may be contrasted with $(A_7,B_5)$, where the change in $A$ is sufficient to distinguish the pair.
Alignment with hard negatives holds $A$ and $S$ fixed while $B$ varies, as in $(A_1,B_3)$ versus $(A_1,B_5)$. Multi-positive InfoNCE treats examples sharing $B$ as positives rather than false negatives.
Full-target latent prediction uses normalized MSE and no negatives.
Each condition uses AdamW with learning rate $3\times10^{-4}$, weight decay $10^{-5}$, batch size 256, 1,500 updates, and pretraining seeds 0--2.

\paragraph{Evaluation.}
After pretraining, the Graph Encoder is frozen.
An $\ell_2$-regularized logistic-regression probe ($C=1$) is fitted on 8,192 generated examples and evaluated on 4,096 disjoint examples. Eight-way chance accuracy is 12.5\%.
For the counterfactual sensitivity check, we replace $B$ with $(B+1)\bmod 8$ while holding $A$ and $S$ fixed, and report $1-\cos(\mathbf g(A,B,S),\mathbf g(A,B',S))$.

\begin{table}[H]
\centering
\caption{Per-seed factor-\(B\) recovery and counterfactual sensitivity in the controlled diagnostic. Probe values are accuracies (\%). Sensitivity is the mean cosine distance after changing only \(B\).}
\label{tab:feature-suppression-diagnostic}
\small
\begin{tabular}{lrrrrr}
\toprule
Condition & Seed 0 & Seed 1 & Seed 2 & Mean & Sensitivity \\
\midrule
Alignment with easy negatives & 12.26 & 13.06 & 12.74 & 12.69 & 0.001 \\
Alignment with hard negatives & 100.00 & 99.83 & 100.00 & 99.94 & 0.835 \\
Full-target latent prediction & 87.45 & 100.00 & 99.41 & 95.62 & 0.535 \\
\bottomrule
\end{tabular}
\end{table}

The results separate two routes to preserving \(B\). Under easy-negative matching, the candidate set lets the model distinguish pairs using the stronger cue \(A\), so \(B\) remains poorly decodable. In the hard-negative condition, holding \(A\) and \(S\) fixed while varying \(B\), together with treating examples that share \(B\) as positives, makes \(B\) relevant to the matching decision. Full-target prediction instead includes \(B\) directly in every target and yields high probe accuracy without factor-specific negative construction. The diagnostic therefore shows how both target content and candidate composition shape which factors remain decodable.

\paragraph{WikiBio two-node diagnostic.}
We use only WikiBio's official training split and leave its validation and test splits untouched. We select examples with one-token nationality and occupation fields from the four nationalities (American, Indian, British, and French) and four occupations (politician, actor, actress, and writer). The first sentence must state the nationality and mention the occupation; examples containing dates or another selected nationality or occupation are excluded, and exact delexicalized sentences are deduplicated. For each of the 16 nationality--occupation cells, seven examples are used for training and two for testing, giving 112 training and 32 held-out examples. The cell-wise split uses seed 20260925.

We construct a two-node connected graph from each selected WikiBio example. The node features are frozen all-MiniLM-L6-v2 embeddings of the nationality and occupation labels, and the target is the same encoder's frozen embedding of the original first-sentence predicate after removing the biography subject. Easy Alignment samples four candidates with distinct nationalities, so occupation is unnecessary for identifying the matching description. Hard Alignment fixes nationality and uses four distinct occupations. Full-target latent prediction uses four sampled graph--description pairs without negatives. All three conditions use a two-dimensional Graph Encoder representation and the same prediction head, batch size 4, 1,500 updates, AdamW learning rate $3\times10^{-4}$, and weight decay $10^{-5}$; the Alignment temperature is 0.07. We train with seeds 0, 1, and 2. After training, we freeze the Graph Encoder and fit a four-way occupation probe using standardized features and logistic regression ($C=1$) on the 112 training examples. Accuracy is measured on the 32 held-out examples; chance accuracy is 25\%. This auxiliary experiment uses a compact two-node model and does not evaluate the complete GL-JEPA architecture.

\section{Model, Pretraining, and Optimization Details}
\label{app:implementation-details}

\begin{table}[H]
\centering
\caption{Default model configuration. The same frozen Text Encoder \(E_T\) embeds node text and paired descriptions.}
\label{tab:model-configuration}
\small
\begin{tabular}{@{}ll@{}}
\toprule
Component & Configuration \\
\midrule
Text Encoder (\(E_T\)) & all-MiniLM-L6-v2 \citep{reimers2019sentencebert}, mean pooling, \(d_T=384\), frozen \\
Graph Encoder (\(E_G\)) & 12 GraphGPS layers, GraphSAGE local operator, \(d_G=384\) \\
Positional encoding & 32-step random-walk features projected to 8 dimensions \\
Predictor (\(P_\omega\)) & final 8 Llama-3.2-1B blocks \citep{meta2024llama32}, layers 8--15, \(d_P=2048\) \\
Predictor input & graph summary followed by contextualized node states \\
Node-token order & ascending original node IDs within each sampled graph \\
\bottomrule
\end{tabular}
\end{table}

\begin{wrapfigure}{r}{0.3\textwidth}
  \centering
  \vspace{-5mm}
  \includegraphics[width=\linewidth]{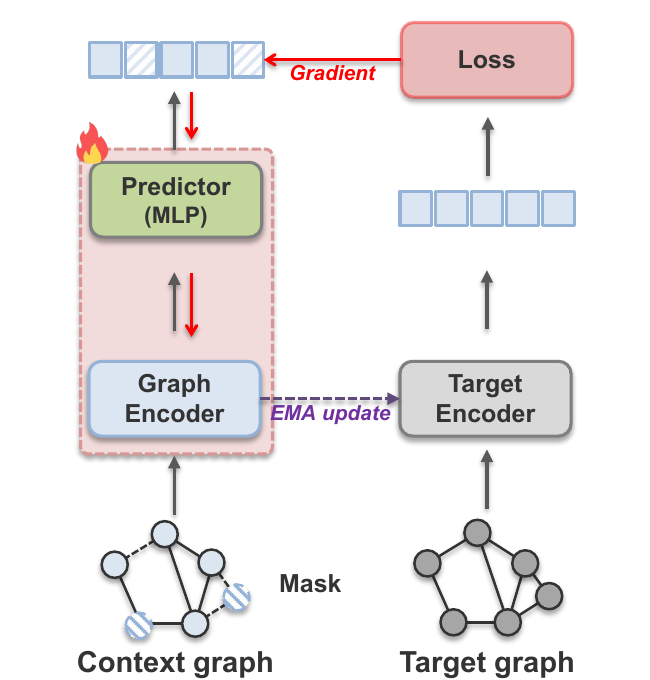}
  \vspace{-2mm}
  \caption{
    Graph Context Pretraining.
  }
  \label{fig:graph-context-pretraining}
    \vspace{-8mm}
\end{wrapfigure}

The graph readout is the affine map
\begin{equation}
 \mathbf g_i=W_g\boldsymbol\mu_i+\mathbf b_g,
 \qquad W_g\in\mathbb R^{d_G\times d_G},\quad \mathbf b_g\in\mathbb R^{d_G},
 \label{eq:app-graph-readout}
\end{equation}
where \(\boldsymbol\mu_i=N_i^{-1}\sum_n\mathbf h_{in}\). This graph summary is used by Graph--Language Latent Prediction.

\paragraph{Graph Context Pretraining.}

The context view masks node features with probability \(0.35\), independently masks feature
coordinates with probability \(0.10\), and drops directed edges with probability \(0.25\).
The target branch receives the unmasked graph. The two-layer structural Predictor maps the
contextualized state of each masked node to its target state. For
\[
h_\beta(\delta)=
\begin{cases}
\delta^2/(2\beta),&|\delta|<\beta,\\
|\delta|-\beta/2,&\text{otherwise},
\end{cases}
\qquad \beta=1,
\]
the Smooth-L1 loss is
\(\ell_{\mathrm{SL1}}(\mathbf a,\mathbf b)=d_G^{-1}\sum_{k=1}^{d_G}h_\beta(a_k-b_k)\).
At update \(t\), the EMA momentum is
\begin{equation}
 \tau_t=0.996+(1.0-0.996)\frac{\min(t,18180)}{18180},\qquad
 \bar\theta\leftarrow\tau_t\bar\theta+(1-\tau_t)\theta.
 \label{eq:app-ema-schedule}
\end{equation}
The transferred checkpoint is taken after 5,454 updates, for which \(\tau_t=0.9972\).

\begin{table}[H]
\centering
\caption{Optimization and capacity of the controlled interfaces without Graph Context Pretraining. Parameter counts are for trainable tensors stored in the evaluated checkpoints.}
\label{tab:controlled-interface-optimization}
\small
\begin{tabular}{lrrrrr}
\toprule
Interface & Updates & Batch & LR & Weight decay & Params (M) \\
\midrule
Token & 18,180 & 512 & \(10^{-5}\) & \(10^{-5}\) & 1,031.2 \\
Alignment & 18,180 & 512 & \(10^{-5}\) & \(10^{-5}\) & 18.5 \\
GL-JEPA & 18,180 & 512 & \(10^{-5}\) & \(10^{-5}\) & 772.6 \\
\bottomrule
\end{tabular}
\end{table}

All three interfaces use AdamW, the same five graph--description corpora, the same
12-layer Graph Encoder architecture, and the same evaluated update count. \textcolor{black}{Table~\ref{tab:frozen-transfer} reports three-pretraining-seed means for all interfaces.} \textcolor{black}{Language-side modules and parameter counts differ across interfaces.} The Linear Predictor in
Table~\ref{tab:frozen-transfer} provides the \textcolor{black}{nearly parameter-matched} control.
Graph--Language Latent Prediction uses normalized MSE with \(1/(Bd_T)\). Both pretraining phases use
\(u_\varepsilon(\mathbf v)=\mathbf v/\max(\lVert\mathbf v\rVert_2,10^{-8})\).

\paragraph{Ablation protocol.}
Each Table~\ref{tab:ablation} variant changes only the component named in its row unless it removes a
pretraining stage. Text Encoder variants use the selected frozen encoder for both node text and paired
descriptions, with mean pooling and \(\ell_2\) normalization. For each encoder, Graph Context
Pretraining is rerun for 5,454 updates, and its Graph Encoder checkpoint initializes 18,180 updates of
Graph--Language Latent Prediction. The remaining data order, optimizer, batch size, and evaluation
contract are fixed across these runs. Predictor-architecture, Predictor-input, and Graph Encoder
variants retain the default settings in Table~\ref{tab:model-configuration} except for the named change.
The completed GraphSAGE variant uses four message-passing layers and is evaluated at step 18,180.

\paragraph{Directed-edge handling.}
The cached edge index preserves source-to-target direction. The GraphSAGE operator
\citep{hamilton2017inductive} aggregates messages along this stored direction. Preprocessing does
not symmetrize the graph or add reverse edges.

\paragraph{Compute and availability.}
Experiments use one NVIDIA RTX PRO 6000 Blackwell GPU (96\,GB), PyTorch 2.11 with CUDA 13.0,
Python~3.12, bfloat16 precision, and AdamW. Training and evaluation code, data manifests, and
checkpoint identifiers will accompany the released artifact.

\section{Interfaces, Decoding, and Instruction Tuning}
\label{app:rq1-protocol}

This section specifies how the learned graph states are used in the controlled comparison,
generation evaluation, valid-description scoring, and graph question answering.

\paragraph{Controlled supervision interfaces.}
Token trains the graph-to-language Llama decoder with teacher-forced next-token cross-entropy.
Alignment compares the pooled graph representation and frozen description representation using
symmetric InfoNCE with temperature \(0.07\). GL-JEPA predicts the frozen description target as
\begin{equation}
 \widehat{\mathbf z}_Y=P_\omega(S_G),\qquad
 \mathbf z_Y=\operatorname{sg}\!\left(E_T(Y)\right).
 \label{eq:app-language-prediction}
\end{equation}
The graph inputs, paired descriptions, Graph Encoder architecture, initialization, batch size, and
evaluated update count are shared. The language-side interfaces remain method-specific.

\paragraph{Generation readout.}
\label{app:text-readout}
For WebNLG generation, the frozen GL-JEPA checkpoint predicts one 384-dimensional description
representation. A trainable adapter \(q_\phi:\mathbb R^{384}\rightarrow\mathbb R^{16\times2048}\)
maps it through a two-layer Transformer to 16 virtual prefix tokens for a frozen
\texttt{meta-llama/Llama-3.2-1B} decoder as
\begin{equation}
 p_\phi(Y\mid G)=\prod_{t=1}^{|Y|}
 p\!\left(y_t\mid y_{<t},q_\phi(\widehat{\mathbf z}_Y)\right).
 \label{eq:app-generation-readout}
\end{equation}
Only \(q_\phi\) is optimized. The adapter is trained with teacher-forced description loss on
cached predicted representations and paired descriptions from the five pretraining corpora in
Table~\ref{tab:pretraining-statistics}, not on WebNLG examples. Training uses 27,270 updates
with batch size 4, gradient accumulation 4, AdamW learning rate \(10^{-4}\), weight decay
\(0.01\), seed 90, and maximum target length 192. A held-out portion of the pretraining
corpora is used for validation.
Test generation is greedy (\texttt{do\_sample=False}) with at most 128 new tokens.

Figure~\ref{fig:dpa-efficiency}(c) reports training-update latency at batch size 8 on an NVIDIA RTX PRO 6000 Blackwell Max-Q. Token includes its Graph Encoder, teacher-forced causal decoder loss, backward pass, and optimizer update with maximum target length 192. GL-JEPA's reported total adds Graph Encoder and Predictor training, backward and optimizer update, and a separately measured teacher-forced generation-readout fitting step through the frozen decoder. Model loading, tokenization, data materialization, validation, logging, and checkpoint writing are excluded. Figure~\ref{fig:dpa-efficiency}(d) reports graph-to-text inference on 64 WebNLG test examples with the same batch size, greedy decoding, and at most 32 generated tokens. Each measurement uses two warm-up batches and five repeats. Component timings use cached outputs from the preceding component. The GL-JEPA training total in panel (c) sums the separately timed pretraining and readout-fitting components, whereas inference totals in panel (d) are measured through the complete method-specific path.

\paragraph{Valid-description scoring.}
For dual-positive accuracy (DPA), Token scores candidates with length-normalized sequence
likelihood, whereas GL-JEPA uses
\begin{equation}
 s_{ij}=\cos\!\left(\widehat{\mathbf z}_{Y_i},E_T(Y_{ij})\right).
 \label{eq:app-dpa-score}
\end{equation}
For two valid descriptions \(Y_{i1},Y_{i2}\) and a negative \(Y_{i-}\),
\begin{equation}
 \operatorname{DPA}=\frac{1}{N}\sum_{i=1}^{N}
 \mathbb 1[s_{i1}>s_{i-}\land s_{i2}>s_{i-}].
 \label{eq:app-dpa}
\end{equation}
Ties count as failures. WebNLG uses 1,744 one-fact counterfactuals comprising 676 entity substitutions,
535 direction reversals, 304 numeric substitutions, and 229 relation substitutions. Each mutation
changes one fact signature while retaining the remaining surface template. No model-based fluency
filter is applied. \textcolor{black}{DPA is the fraction of examples where both valid descriptions score above the
constructed negative.} DART uses 3,621 examples and pairs two valid references with a description
from another graph.

\begin{table}[H]
\centering
\caption{Decomposition of DPA (\%). R1 and R2 denote the rates at which each valid reference outranks the negative. DPA requires both comparisons to succeed.}
\label{tab:dpa-decomposition}
\small
\begin{tabular}{llrrrr}
\toprule
Dataset & Model & Examples & R1 & R2 & DPA \\
\midrule
WebNLG & Token & 1,744 & \textcolor{lowergray}{51.78} & \textcolor{lowergray}{55.85} & \textcolor{lowergray}{35.55} \\
\rowcolor{gljepablue}WebNLG & \textbf{GL-JEPA} & 1,744 & 69.67 & 76.78 & 60.78 \\
DART & Token & 3,621 & \textcolor{lowergray}{50.51} & \textcolor{lowergray}{51.73} & \textcolor{lowergray}{39.82} \\
\rowcolor{gljepablue}DART & \textbf{GL-JEPA} & 3,621 & 88.18 & 87.90 & 83.57 \\
\bottomrule
\end{tabular}
\end{table}

\paragraph{Query-conditioned instruction tuning.}
\label{app:query-conditioned-qa}
For each example, the valid sequence is
\[
[\texttt{graph-summary},\texttt{node}_1,\ldots,\texttt{node}_N,
 \texttt{query}_1,\ldots,\texttt{query}_K].
\]

Query tokens use the Llama-3.2-1B tokenizer and input embedding matrix followed by a learned
projection to \(d_P\). Rotary positions continue across graph, node, and query tokens. Bidirectional
attention is enabled between every valid position and masks only padding. The readout averages valid
query-token outputs. The Graph Encoder and frozen answer Text Encoder \(E_A\) are fixed, and only
the query-conditioned Predictor and output projection are adapted as
\begin{equation}
 \widehat{\mathbf z}_A=P_\omega^{\mathrm{task}}(S_G,Q),\qquad
 s(a\mid G,Q)=\cos\!\left(\widehat{\mathbf z}_A,E_A(a)\right).
 \label{eq:app-query-adaptation}
\end{equation}
\begin{wrapfigure}{r}{0.32\textwidth}
\centering
\vspace{-0.8\baselineskip}
\includegraphics[width=\linewidth]{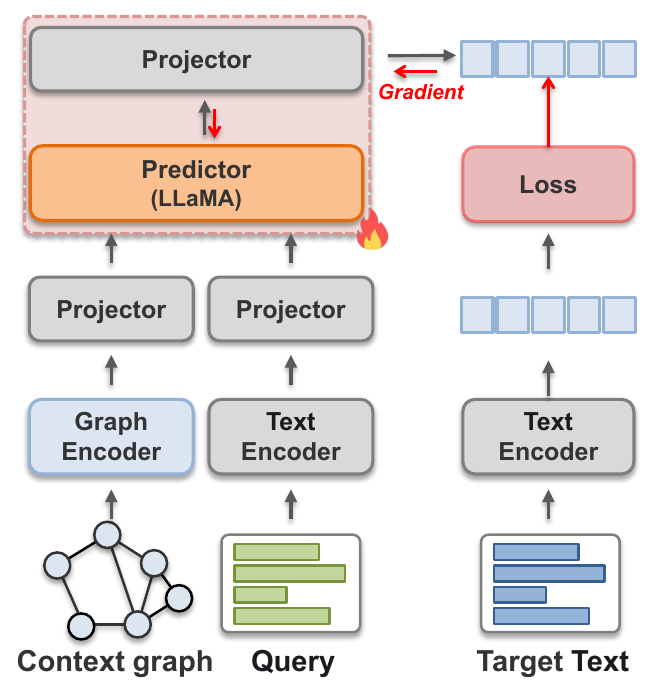}
\caption{Query-conditioned adaptation. The Graph Encoder and answer Text Encoder \(E_A\) are frozen. The query-conditioned Predictor and output projection are adapted.}
\label{fig:query-conditioned-gqa}
\vspace{-0.5\baselineskip}
\end{wrapfigure}
The default configuration uses the same encoder checkpoint for \(E_A\) as for \(E_T\). Encoder
variants use their corresponding frozen Text Encoder. Candidate embeddings and predictions are
\(\ell_2\)-normalized, and candidate cross-entropy uses temperature \(0.07\).

Following the discriminative VQA evaluation used by VL-JEPA \citep{chen2025vljepa}, we score the
deterministic candidate set \(\{\texttt{no},\texttt{yes}\}\) against the query-conditioned
prediction and select the higher-scoring answer. Training contains 11,433 examples, validation
contains 1,267, and test contains 400 examples per task. Adaptation uses five epochs, batch size 8,
candidate batch size 32, AdamW learning rate \(10^{-5}\), weight decay \(0.05\), and seed 90.
Validation accuracy selects the checkpoint.

\end{document}